\documentclass{article}

\PassOptionsToPackage{numbers, compress}{natbib}
\usepackage[preprint]{neurips_2025}

\usepackage[utf8]{inputenc} % allow utf-8 input
\usepackage[T1]{fontenc}    % use 8-bit T1 fonts
\usepackage{hyperref}       % hyperlinks
\usepackage{url}            % simple URL typesetting
\usepackage{booktabs}       % professional-quality tables
\usepackage{amsfonts}       % blackboard math symbols
\usepackage{nicefrac}       % compact symbols for 1/2, etc.
\usepackage{microtype}      % microtypography
\usepackage{xcolor}         % colors

\usepackage{amsmath}
\usepackage{amssymb}
\usepackage{mathtools}
\usepackage{amsthm}

\usepackage{booktabs}       %
\usepackage{multirow}
\usepackage{multicol}

\usepackage[capitalize,noabbrev]{cleveref}

\theoremstyle{plain}

\theoremstyle{definition}

\theoremstyle{remark}

\title{Interpretable Zero-shot Learning \\ with Infinite Class Concepts}

\author{%
  Zihan Ye \\
  Xian Jiaotong-Liverpool University\\
  \texttt{zihhye@outlook.com} \\
  \And
  Shreyank N Gowda \\
  University of Nottingham\\
  \texttt{kini5gowda@gmail.com} \\
  \And
  Shiming Chen \\
  Mohamed bin Zayed University of Artificial Intelligence\\
  \texttt{gchenshiming@gmail.com} \\
  \And
  Yaochu Jin \\
  Westlake University\\
  \texttt{jinyaochu@westlake.edu.cn} \\
  \And
  Kaizhu Huang \\
  Duke Kunshan University\\
  \texttt{kaizhu.huang@dukekunshan.edu.cn} \\
  \And
  Xiaobo Jin \\
  Xian Jiaotong-Liverpool University\\
  \texttt{Xiaobo.Jin@xjtlu.edu.cn} \\
}

\begin{document}

\maketitle

\begin{abstract}
Zero-shot learning (ZSL) aims to recognize unseen classes by aligning images with intermediate class semantics, like human-annotated concepts or class definitions. An emerging alternative leverages Large-scale Language Models (LLMs) to automatically generate class documents. However, these methods often face challenges with transparency in the classification process and may suffer from the notorious hallucination problem in LLMs, resulting in non-visual class semantics. This paper redefines class semantics in ZSL with a focus on transferability and discriminability, introducing a novel framework called Zero-shot Learning with Infinite Class Concepts (InfZSL). Our approach leverages the powerful capabilities of LLMs to dynamically generate an unlimited array of phrase-level class concepts. To address the hallucination challenge, we introduce an entropy-based scoring process that incorporates a ``goodness" concept selection mechanism, ensuring that only the most transferable and discriminative concepts are selected. Our InfZSL framework not only demonstrates significant improvements on three popular benchmark datasets but also generates highly interpretable, image-grounded concepts.
Code will be released upon acceptance.
\end{abstract}

\section{Introduction}
\label{sec:intro}

\begin{figure*}[htbp]
    \centering
    \includegraphics[width=0.9\linewidth]{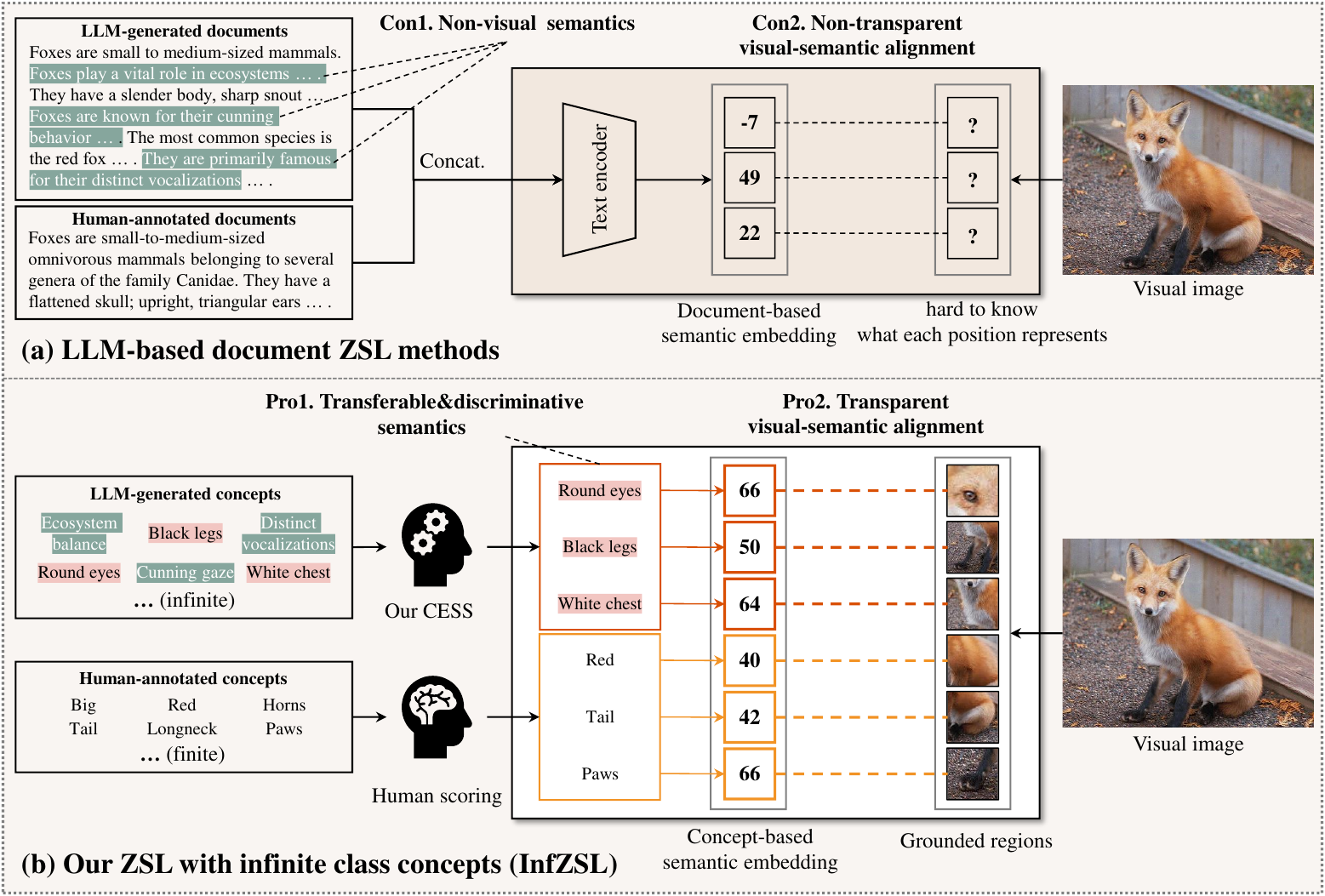}
    \caption{  Motivation Illustration. (a) LLM-based document ZSL methods encounter two main issues: non-visual semantics caused by LLM hallucination and a lack of transparency in visual-semantic alignment. (b) Our InfZSL addresses these challenges by introducing Concept Entropy Selection and Scoring (CESS), which selects and scores concepts with high transferability and discriminability. InfZSL also enables transparent visual-semantic alignment, enhancing interpretability in the ZSL decision-making process.}
    \label{fig:banner}
    \vspace{-0.2cm}
\end{figure*}

Human learning involves a remarkable ability to imagine and recognize unseen objects from descriptions alone~\cite{pearson2019human}.
Equipping machines with similar capabilities could greatly reduce costs associated with data collection and model training.
In computer vision, this challenge is addressed through Zero-Shot Learning (ZSL), which enables models to predict unseen classes by linking images with intermediate class semantics.
Existing approaches typically rely on human-annotated documents~\cite{naeem2022i2dformer} and concepts~\cite{hou2024visual}.
However, creating  annotations at scale is costly and requires domain expertise~\cite{yu2013designing, song2018selective}.
Consequently, many works have focused on automatic methods for semantic mining~\cite{naeem2022i2dformer,xu2022vgse}.

Inspired by the impressive capabilities of Large Language Models (LLMs)~\cite{achiam2023gpt,team2023gemini}, 
recent approaches have attempted to automate the generation of class documents~\cite{naeem2023i2mvformer, qu2024visual}.
These methods combine multiple LLM-generated documents with human-annotated sources (e.g., Wikipedia) to compile comprehensive class semantics. 
By concatenating all documents and feeding them into a text encoder, models can obtain a semantic embedding that aligns with visual data.

Despite substantial progress in LLM-based document ZSL (see Fig.~\ref{fig:banner} (a)), two critical challenges persist:
\begin{enumerate}
    \item \textbf{Non-visual semantics.} LLMs are prone to generating irrelevant content due to the well-documented 
     tendency—an issue commonly referred to as the hallucination problem~\cite{ji2023survey, maynez2020faithfulness}.
    For example, explicitly defined prompts that specify visual image-related semantics often yield irrelevant outputs such as ``vital role in ecosystems," ``cunning behavior," or ``distinct vocalization," which are difficult to connect to visual features~\cite{shang2024incremental}.
    Such irrelevant semantics impair the transfer of visual knowledge to recognize unseen classes.
    \item \textbf{Non-transparent visual-semantic alignment.} Although document-based semantic embeddings can be generated, the black-box nature of text and visual encoders makes interpretation challenging~\cite{castelvecchi2016can}. The specific significance of each position within the embedding remains elusive, obscuring insights into the decision-making mechanisms underpinning ZSL. This non-transparency can lead to the inadvertent incorporation of extraneous semantics into the embedding, culminating in unwarrantable ZSL outcomes.

    % We do not know what each position in the embedding represent, thus hindering our understanding of the ZSL decision-making process.
    % This could result in irrelevant semantics being mixed into the semantic embedding, resulting in unwarrantable ZSL.
\end{enumerate}

To address the mentioned challenges, we pivot towards concept-based methods enjoying a more transparent ZSL decision process than document-based approaches.
Specifically, they required manually defined concept sets, which in turn had to be scored by experts to craft semantic embeddings for each category, but the key bottlenecks are that the human-annotated concept is `finite', and the annotation still is expensive. Thus, the crucial question emerges for the current ZSL community:

\textit{Could concept-based ZSL methods also embrace the advent of powerful LLMs to automatically obtain 'infinite' class concepts from LLMs and utilize in the full ZSL pipeline?} 

To answer the question, we introduce \textbf{Zero-Shot Learning with Infinite Class Concepts (InfZSL)}, which automates concept generation, selection, and scoring grounded on a set of well-defined criteria (Fig.~\ref{fig:banner} (b)).
Our method focuses on generating infinite, LLM-derived class concepts using carefully crafted prompts, followed by filtering and scoring concepts based on two essential factors: \textbf{transferability} and \textbf{discriminability}.
Specifically, we define a new metric, called \emph{concept entropy}.
It allows us to measure and select concepts that are both highly transferable across classes and discriminative enough to separate different categories. Distinct with traditional semantic entropy methods~\cite{farquhar2024detecting}, which detect hallucination at the level of entire generated content, our concept entropy can detect hallucination within individual concepts. Moreover, it quantifies not only unfaithful concepts but also those that lack sufficient transferability or discriminability.
Furthermore, we propose the concept-entropy-based selection and scoring (CESS) strategy to mitigate hallucinated concepts and explicitly score them according to the class-concept correlation.
Finally, we can easily integrate the infinite generated concepts with existing concept-based methods.
In summary, our InfZSL approach provides a transparent and interpretable decision-making process by representing each embedding position with human-understandable, phrase-level concepts that can be visualized in images.
This transparent alignment between visual data and semantic concepts sets a new standard for interpretable ZSL. 

We summarize our contributions as follows. 
\begin{enumerate}
    \item  We provide a novel framework that supplements finite human-annotated class concepts with infinite LLM-generated class concepts.
    
    \item We delve into the hallucination problem in the ZSL task, and propose a novel concept entropy, eliminating hallucinated concepts as well as selecting those concepts that share both high transferability and discriminability.
    % \item Our model InfZSL consistently improves SOTAs across various datasets and methods.
    
    \item We qualitatively demonstrate that our method not only improves accuracy over SOTAs across various datasets and methods, but also helps with interpretablity in ZSL.
\end{enumerate}

\section{Related work}
\label{sec:relatedwork}

%-------------------------------------------------------------------------
\subsection{Zero-shot Learning}
Zero-shot learning (ZSL) addresses the generalization challenge of transferring a model trained on seen classes to make predictions on unseen classes \cite{xian2018zero}.
This approach relies on aligning images of seen classes with shared class semantics that can generalize to unseen classes.

ZSL methods are generally categorized into embedding and generative approaches.
Embedding methods learn the similarity between images and class semantic embeddings directly \cite{ye2023rebalanced, chen2022transzero++, chen2024progressive}.
However, due to the lack of samples from unseen classes, these methods often produce predictions that are biased towards seen classes \cite{min2020domain}.
Generative methods, on the other hand, leverage generative models (e.g. GAN, VAE) to learn feature generation based on class semantics from seen classes.
Once trained, these models can generate features for unseen classes, which are then used to train a final ZSL classifier with pseudo-features.
Most of these methods, however, assume that class semantics are manually annotated, which is costly, difficult to scale for large datasets, and cannot fully capture the diversity of all semantics \cite{rohrbach2017generating, zhou2019grounded}.
Consequently, developing automated methods for semantic annotation has become an urgent need.

%-------------------------------------------------------------------------
\subsection{Automated Semantic Annotation}

Automatic semantic annotation aims to obtain class semantics without human intervention.
In the ZSL field, early works relied on embedding class names using word2vec \cite{mikolov2013distributed} or TF-IDF \cite{salton1988term, qiao2016less}, which represent class documents based on word frequencies.
Subsequent works introduced more sophisticated approaches. VGSE \cite{xu2022vgse} first learns semantics from image patches of seen classes and then extrapolates unseen class semantics based on similarities between seen and unseen class names. I2DFormer \cite{naeem2022i2dformer} uses a two-branch transformer to project single-class documents and class images into a shared semantic space. I2MVFormer \cite{naeem2023i2mvformer} goes further by incorporating multiple class documents to create a more comprehensive semantic representation. Beyond ZSL, Concept Bottleneck Models (CBMs) also seek to automate class semantics but do not differentiate between seen and unseen classes. LaBo \cite{yang2023language} uses LLM-generated concepts, which are then scored by the pretrained vision-language model CLIP \cite{radford2021learning}. Res-CBM \cite{shang2024incremental} introduces a concept discovery module that incrementally identifies potential concepts to complete class semantics. Our work differs from these approaches in four key ways: (1) most existing ZSL and CBM methods do not address the hallucination problem in LLMs; (2) existing ZSL methods rely solely on LLM-generated document semantics or unsupervised embeddings, lacking interpretable semantic embeddings; and (3) CBMs do not consider the selection and scoring of concepts specifically for unseen classes.

\subsection{Large Language Models}
Large Language Models (LLMs), such as ChatGPT \cite{achiam2023gpt} and Gemini \cite{team2023gemini}, are trained on massive web-scale datasets. These models exhibit impressive capabilities across a wide range of tasks, including reasoning \cite{wei2022chain}, question answering \cite{kamalloo2023evaluating}, and document summarization \cite{zhang2024benchmarking}. However, they often produce unsubstantiated answers or responses lacking necessary information—a phenomenon known as the ``hallucination'' problem~\cite{ji2023survey}. Previous approaches to mitigate hallucination have used supervised truthfulness reinforcement~\cite{xiao2021hallucination,filippova2020controlled,schulman2023reinforcement} or entropy-based uncertainty estimation~\cite{farquhar2024detecting}. For instance, semantic entropy \cite{farquhar2024detecting} is a general method for detecting incorrect answers. This approach generates multiple answers to each question, clusters responses with similar meanings, and calculates entropy across these clusters. Ultimately, it discards responses with high entropy values, indicating lower confidence or coherence.

Our work advances the use of LLMs in two key ways: (1) While existing hallucination detection methods primarily focus on identifying unfaithful answers, our approach recognizes that even accurate class concepts may lack the necessary transferability and discriminability needed. Therefore, our method not only filters out unfaithful concepts but also evaluates each concept’s transferability and discriminability. (2) Additionally, our approach can detect hallucination within partial responses, further enhancing accuracy and reliability.

\section{Methodology}
\label{sec:method}

\begin{figure*}[htbp]
    \centering
    \includegraphics[width=0.9\linewidth]{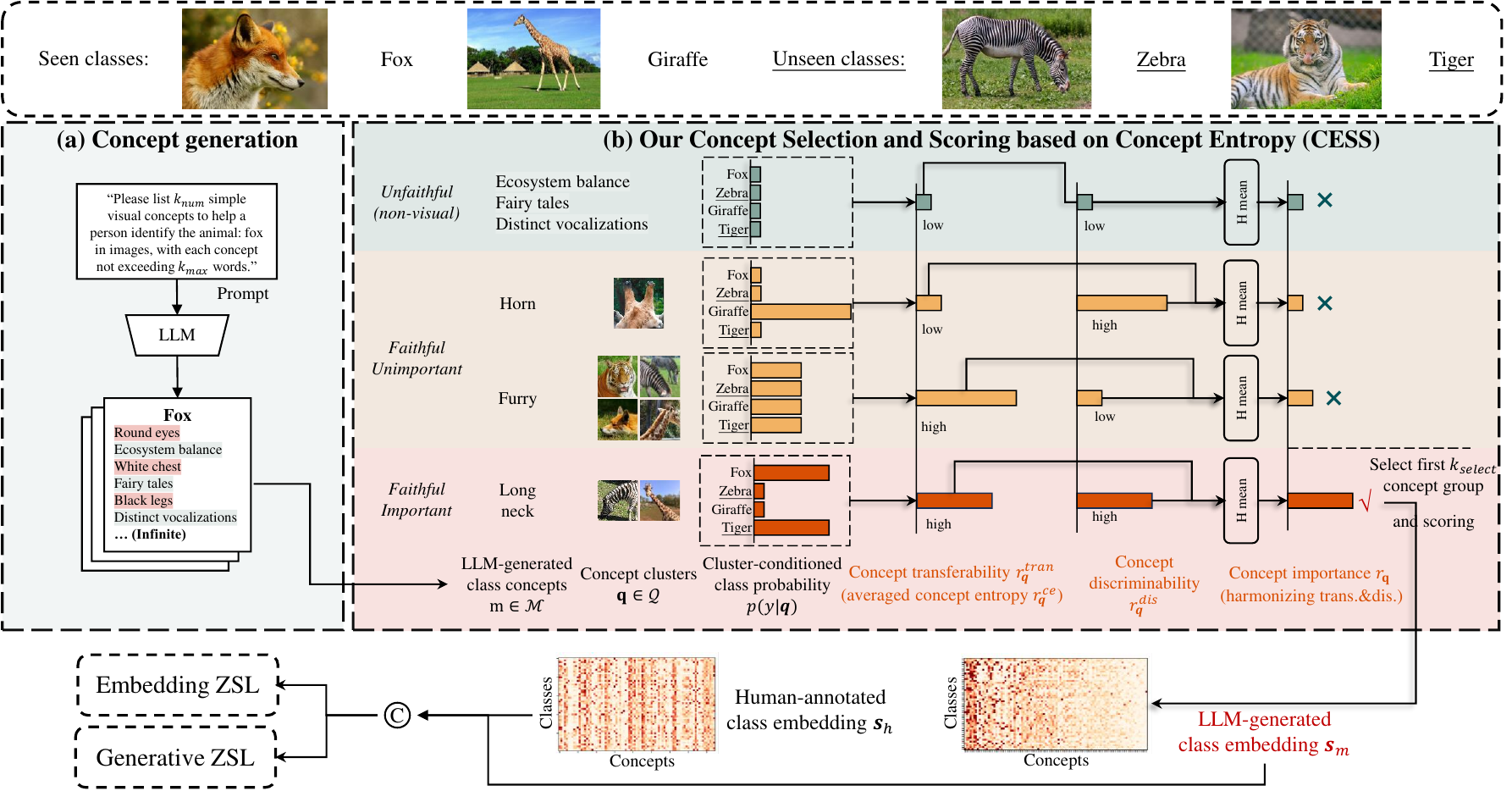}
    \caption{An illustration of our InfZSL. It consists of three steps. (a) Concept generation: we use our designed promt to extract any number of class concepts. (b) Concept selection and scoring: LLMs might generate non-visual concepts; however, even among visual concepts, some may possess only transferability (e.g. `furry' in the illustration.) or discriminative power (e.g. `horn').
    Only the concepts that have both high discriminability and transferability are we need in ZSL (e.g. `long neck').
    We leverage our proposed concept entropy to select and score them to get the class embedding $\mathbf{s}_{m}$ based on LLM-generated concepts.
    (3) Once construct our $\mathbf{s}_{m}$, we can immediately integrate it with human-annotated class embedding $\mathbf{s}_{h}$ into existing concept-based embedding or generative ZSL methods.
    }
    \label{fig:concept_type}
\end{figure*}

% \begin{figure*}[htbp]
%     \centering
%     \includegraphics[width=\linewidth]{infzsl.pdf}
%     \caption{The illustration of our InfZSL.}
%     \label{fig:infzsl}
% \end{figure*}

\begin{figure*}[htbp]
    \centering
    \includegraphics[width=0.85\linewidth]{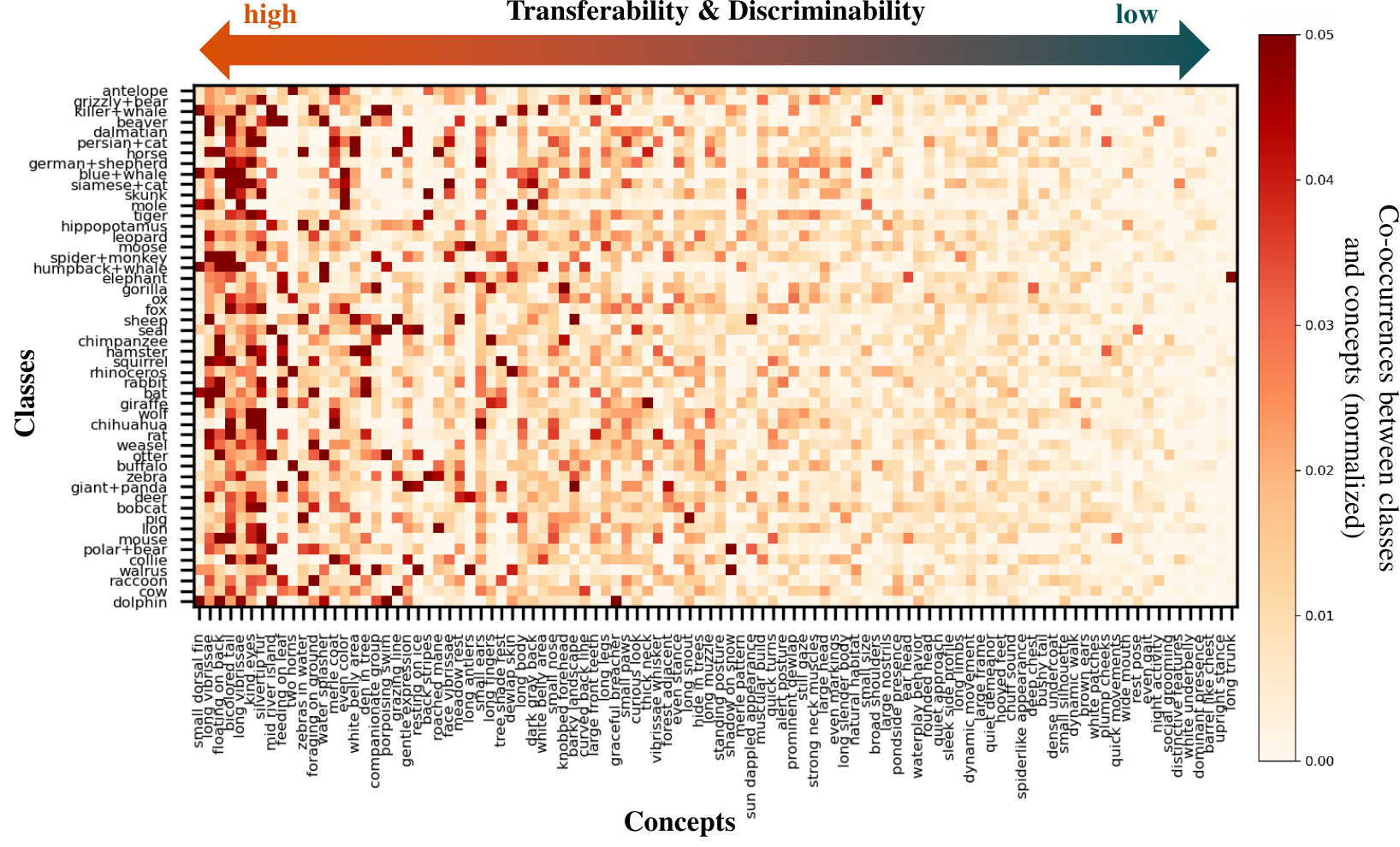}
    \vspace{-0.2cm}
    \caption{The heatmap visualization of our LLM-concept-based semantic embeddings.}
    \vspace{-0.3cm}
    \label{fig:infzsl}
\end{figure*}

Our InfZSL framework involves three stages as shown in Fig.~\ref{fig:infzsl}: (1) Concept generation: we employ a specially designed prompt to generate an infinite set of class concepts from LLMs, enriching the limited set of human-annotated concepts. (2) Concept selection and scoring: to identify the most essential class concepts, we introduce the concept entropy metric, which evaluates each concept’s transferability and discriminability and select them. We score selected concepts by their class-concept co-occurrences to build the class semantic embedding. (3) Concept learning: we integrate this constructed concept semantic embedding with human-annotated embeddings and establish visual-semantic alignment.

\subsection{Problem Formulation}
In ZSL, we denote the set of seen classes as $\mathcal{Y}^{s}$ and the set of unseen classes as $\mathcal{Y}^{u}$, where $\mathcal{Y}^{s} \cap \mathcal{Y}^{u} = \emptyset$. In existing concept-based methods, a set of high-quality concepts $\mathcal{H}$ (e.g., `tails,' `long leg') for all classes is defined by experts~\cite{yang2023language}. Each concept is scored individually for each class, typically based on the number of occurrences~\cite{wah2011caltech}, to obtain the concept-based semantic embedding spaces $\mathcal{S}^{s}_{h}$ for seen classes and $\mathcal{S}^{u}_{h}$ for unseen classes.

Combining $\mathcal{S}^{s}_{h}$ with the images \( \mathcal{X}^{s} \) and labels \( \mathcal{Y}^{s} \) of the seen classes, the training set is constructed as $\mathcal{D}^{tr} = \{(\mathbf{x}^{s}, y^{s}, \mathbf{s}^{s}_{h}) \mid \mathbf{x}^{s} \in \mathcal{X}^{s}, y^{s} \in \mathcal{Y}^{s}, \mathbf{s}^{s}_{h} \in \mathcal{S}^{s}_{h}\}$. The objective in ZSL is to use this training dataset \( \mathcal{D}^{tr} \) to create a classifier capable of predicting unseen classes for images in the test dataset \( \mathcal{D}^{te} = \mathcal{D}^{u} = \{(\mathbf{x}^{u}, y^{u}, \mathbf{s}^{u}_{h}) \mid \mathbf{x}^{u} \in \mathcal{X}^{u}, y^{u} \in \mathcal{Y}^{u}, \mathbf{s}^{s}_{h} \in \mathcal{S}^{s}_{h} \} \), i.e., \( f_{zsl}: \mathcal{X}^{u} \to \mathcal{Y}^{u} \).

In the Generalized ZSL (GZSL) task, test samples may come from both seen and unseen classes. Let $\mathcal{D}^{te,s}$ represent the portion of seen class samples reserved for testing, so the testing dataset becomes \( \mathcal{D}^{te} = \mathcal{D}^{te,s} \cup \mathcal{D}^{u} \). The goal in GZSL is then defined as \( f_{gzsl}: \mathcal{X}^{s} \cup \mathcal{X}^{u} \to \mathcal{Y}^{s} \cup \mathcal{Y}^{u} \).

\subsection{Concept Generation}
Manually annotating concepts is challenging to scale, costly, and makes it difficult to cover all possible concepts. To address this, we leverage LLMs to generate class concepts to enrich human-annotated ones. Following previous LLM-based ZSL work~\cite{naeem2023i2mvformer} and CBM work~\cite{yang2023language}, we designed the following prompt template:

\textit{
``Please list $\{k_{num}\}$ simple visual concepts to help a person identify the $\{class\_type\}$: $\{class\_name\}$ in images, with each concept not exceeding $\{k_{max}\}$ words."
}

The template arguments can be filled in easily to prompt the LLMs. Here, $\{class\_name\}$ is the specific class name, and $\{class\_type\}$ denotes the category, such as `animals' for AWA2~\cite{xian2018zero}, `birds' for CUB~\cite{wah2011caltech}, and `scenes' for SUN~\cite{patterson2012sun}. We also define $k_{num}$ and $k_{max}$ to control the number of generated concepts and the maximum word count per concept, respectively. For $k_{max}$, we recommend a small value (e.g., 1–5), as longer concepts can hinder transferability~\cite{shang2024incremental}. 

While $k_{num}$ can theoretically be set to any value, we observed that excessively large values exacerbate the hallucination problem. Therefore, we set $k_{num}$ to 100 and prompt $k_{time}$ times per class. In other words, the total number of LLM-generated concepts collected per class is $k_{con} = k_{num} \times k_{time}$.

% More details are shown in the supplementary.

\subsection{Concept Selection}

Now, we have a large set, potentially infinite, of class concepts generated by LLMs, denoted as $\mathbf{m} \in \mathcal{M}$.
Different LLMs-generated concepts might share similar meanings, e.g.,  `hairy' and `furry'.
Thus, we assign the unselected $\mathcal{M}$ into $k_{pre}$ clusters $\mathbf{q} \in \mathcal{Q}$ according to their concept embedding $\mathbf{e}_{m} \in \mathcal{E}_{m}$ extracted from the word representation model GloVe~\cite{pennington2014glove}~\footnote{
Previous CBMs often leverage CLIP~\cite{radford2021learning} to encode concepts~\cite{yang2023language, shang2024incremental}.
However, since CLIP is trained by aligning huge image-text pairs from web, the visual information of unseen classes might be leaked, breakin the zero-shot premise.
Thus, we follow existing ZSL work and use GloVe to avoid the class overlap.
}.
For single-word concepts, we directly use the output of Glove.
For multi-word concepts, we use the mean of the words in the concept.
The cluster algorithm is the classical k-means~\cite{arthur2006k}.

However, these extracted concepts (and clusters) still cannot be directly applied to ZSL because (1) LLMs can produce non-visual concepts due to hallucination, and (2) even among visual concepts, some may excel in only transferability or discriminability.
To select essential concepts, we propose a concept selection strategy.
We introduce it with a simplified example, as shown in Fig.~\ref{fig:concept_type}.
% Here we present how to quantitatively measure these two scores.

Let's assume that the seen classes are fox and giraffe, the unseen classes are zebra and tiger, and LLMs-generated concepts could be clustered as five clusters: (1) `sharp head' for fox, (2) `horn' for giraffe, (3) `furry' for all classes, (4) `long neck' for zebra and giraffe, (5) and `claw' for `fox' and `tiger'.

\subsubsection{Concept Transferability}
A transferable concept should help bridge the gap across classes. In other words, a transferable concept should appear in both some seen and unseen classes, allowing us to transfer learned concepts from seen classes to unseen classes through the prediction of transferable concepts.

For example, the generated concept `furry' is a faithful and visual description, but it is not helpful for recognizing unseen classes since all classes share this characteristic. Thus, the model cannot distinguish between classes using this concept alone.

To quantitatively measure concept transferability $r^{tran}_{\mathbf{q}}$, we also define the new metric \textbf{Concept Entropy} $r^{ce}_{\mathbf{q}}$.
Specifically, we then calculate the class-cluster co-occurrence percent $o_{i,j} \in \mathcal{O}$ as the cluster-conditioned class probability.
For every cluster $\mathbf{q}$:
%\begin{align}
    %p(y|\mathbf{q}_{j}) = \frac{ \exp(o_{y,j})}{ \sum_{ 1 \leq j^{\prime} \leq k_{pre} } \exp(o_{y,j^{\prime}}) }.
%\end{align}
\begin{align}
    p(y|\mathbf{q}) = \frac{ \exp(o_{y,\mathbf{q}})}{ \sum_{ 1 \leq y^{\prime} \leq ||\mathcal{Y}|| } \exp(o_{y^{\prime},\mathbf{q}}) },
\end{align}
where $\mathcal{Y} = \mathcal{Y}^s \cup \mathcal{Y}^u$.
This probability measures the class uncertainty of when a concept (its cluster) appears.
Then, we obtain our concept entropy $r^{ce}_{\mathbf{q}}$ as sum of the entropy of concept-conditioned class probabilities:
\begin{align}
    r^{ce}_{\mathbf{q}}= \sum_{y \in \mathcal{Y}} -p(y|\mathbf{q}) \log p(y|\mathbf{q}).
\end{align}

Next, by normalizing it, we can obtain our transferability metric 
\begin{equation}
    r^{tran}_{\mathbf{q}} = \frac{r^{ce}_{\mathbf{q}}}{\sum_{\mathbf{q}^{\prime}\in \mathcal{Q}} r^{ce}_{\mathbf{q}^{\prime}}}.
\end{equation}

\subsubsection{Concept Discriminability}
A discriminative concept should effectively distinguish between different categories.
In other words, when a discriminative concept appears in an image, it should strongly indicate that the image belongs to a limited set of classes.

To this end, we sort $p(y|\mathbf{q})$ by the descending order among all classes.
Then we use the $k_{top}$-th probability from sorted $\grave{p}$ as our discriminability metric
\begin{equation}
\label{eq:dis}
    r^{dis}_{\mathbf{q}} = \grave{p}(y|\mathbf{q})_{(k_{top})}.
\end{equation}
We use the $k_{top}$-th largest $\grave{p}$ is due to it can considers concepts to more classes.
If we only consider the largest one, the concepts might have small probabilities on other all classes that is conflicting to our concept transferability.

% \begin{equation}
% \label{eq:dis}
%     r^{dis}_{\mathbf{q}} = \min(\text{top-}k_{top}(p(y|\mathbf{q}))).
% \end{equation}

\subsubsection{Harmonizing Selection}
The model should also avoid overly-transferable or overly-discriminative concepts.
For instance, among these four classes, only giraffes have `horns'. When the concept `horn' appears in an image, we can immediately conclude that the image is of a giraffe. However, since no other class shares this concept, `horn' is ineffective for recognizing other classes.

Ideal concepts should have both high transferability and high discriminability.
For example, both giraffes and zebras have `long necks,' meaning that when this concept appears, the model can infer that the image is likely of a giraffe or a zebra.
Thus, our concept importance $r_{\mathbf{q}}$ is the harmonic mean of transferability degree and discriminative degree $r_{\mathbf{q}} = 2\times r^{tran}_\mathbf{q}\times r^{dis}_{\mathbf{q}}/(r^{tran}_\mathbf{q}+r^{dis}_\mathbf{q})$.
We select the concepts from the first $k_{select}$ clusters.
We denote the selected concept clusters as $\mathcal{Q}^{\prime}$.

\subsection{Concept Scoring}
To score our selected concepts clusters $\mathcal{Q}^{\prime}$, we refer to existing human-annotating concept work~\cite{xian2018feature, wah2011caltech} who employ class-concept co-occurrences as concepts semantic embeddings.
Specifically, they leverage human experts to mark concepts is existing or not in class images and average the markings.
Following them, we also count the number of occurrences of selected concept clusters $\mathcal{Q}^{\prime}$ for every classes and use the mean value along classes, resulting our our concept-based semantic embeddings $\mathcal{S}_{m}$.

\subsection{Concept Learning}
Now, we have two concepts sets, one is human-annotated $\mathcal{H}$ and the other is $\mathcal{Q}^{\prime}$ selected from LLM-generated infinite concepts, and their corresponding semantic embeddings are $\mathcal{S}_{h}$ and $\mathcal{S}_{m}$.
we can easily and seamlessly integrate these two class embeddings into existing concept-based embedding methods and generative methods by simply concatenating $\mathcal{S}_{h}$ and $\mathcal{S}_{m}$ as the final class embedding $\mathcal{S}$.
For implementation, we also choose a SOTA generative method ZeroDiff~\cite{ye2024exploring} and design a embedding method I2CFormer.
More implementation details are provided in \textbf{Appendix ~\ref{app:I2CFormer}}.

\section{Experiments}
\label{sec:experiments}
% To demonstrate the effectiveness of our InfZSL, we evaluate our InfZSL on multiple metrics both on ZSL and GZSL settings over three popular benchmark datasets.
% With extensive studies, we show our InfZSL could achieve the SOTA with a significant gap as seen in Sec.~\ref{sec:sota}.
% We also present the ablation study for components and hyper-parameters effectiveness of our infinite concepts in Sec.~\ref{sec:ablation}.
% Finally, we present the visualization about our distilled concepts and attention maps on Sec.~\ref{sec:vis}.

\textbf{Datasets.}
To demonstrate the effectiveness of our InfZSL, we evaluate our InfZSL in three popular ZSL benchmarks: (1) The AWA2~\cite{xian2018zero} with 50 animal classes and 85 human-annotated concepts; (2) A bird dataset CUB~\cite{wah2011caltech} that contains 200 classes with 312 human-annotated concepts; (3) A scene datasets SUN~\cite{patterson2012sun} including 717 classes and 102 human-annotated concepts.

\textbf{Evaluation Prototype.}
Following~\cite{chen2024progressive}, we measure the top-1 accuracy both in the ZSL and GZSL settings.
For ZSL, we calculate the top-1 classification accuracy ($T1$) for unseen classes.
For GZSL, we calculate three kinds of top-1 accuracies, namely the 463
accuracy for unseen classes ($U$), the accuracy for seen classes ($S$), and their harmonic mean $H = (2 \times S \times U)/(S + U)$.
% Besides, we also report the Area Under Seen-Unseen accuracy Curve (AUSUC)~\cite{chao2016empirical} in \textbf{Appendix}, which evaluates the degree of trade-off between $U$ and $S$.

\textbf{Implementation Details.}
Our InfZSL can be intergrated into generative methods and embedding methods.
We choose the ZeroDiff~\cite{ye2024exploring} as our generative baseline and we design I2Cformer as our embedding baseline.
These two are pre-trained on ImageNet-1k~\cite{deng2009imagenet} for fair comparison.
We use the class splitting proposed in~\cite{xian2018zero} that ensures the test classes excluded from ImageNet-1k.
For concept generation, we empirically set $k_{num}$ and $k_{max}$ to 100 and 3 for all datasets.
We set $k_{time}$ to 5 for two datasets AWA2 and CUB, but set $k_{time}$ to 1 for SUN as their different class numbers.
For concept selection, we empirically set the hyper-parameters ($k_{pre}$, $k_{select}$, $k_{top}$) to (200, 60, 3), (500, 200, 10) and (200, 100, 10) for AWA2, CUB and SUN, respectively.
%We perform experiments on a single NVIDIA Tesla 3090 graphic card with 24GB memory.
%We use PyTorch for the implementation of all experiments.

\subsection{Comparing with State-of-the-Art}
\label{sec:sota}
We compare our method to concept-based embedding and generative methods.
Our approach with embedding methods: APN~\cite{xu2020attribute}, TransZero~\cite{chen2022transzero}, DUET~\cite{chen2023duet} and ZSLViT~\cite{chen2024progressive}, and generative methods: HSVA~\cite{chen2021hsva}, DSP~\cite{chen2023evolving} VADS~\cite{hou2024visual} and ZeroDiff~\cite{ye2024exploring}.
Our main counterparts also include the document-based methods I2DFormer~\cite{naeem2022i2dformer} and I2MVFormer~\cite{naeem2023i2mvformer}.

% \textbf{Comparing with ZSL SOTA.}

\begin{table*}[htbp]
\setlength\tabcolsep{2pt}
	\centering
	\caption{Comparisons with the state-of-the-arts. For ZSL, T1 denotes  the top-1 accuracy (\%) of unseen classes. For GZSL, \( U \), \( S \), and \( H \) represent the top-1 accuracy (\%) of unseen classes, seen classes, and their harmonic mean, respectively.
    The type `E' and `G' denotes embedding  and generative ZSL methods, respectively.
    The symbol $\dagger$ denotes concept-based ZSL methods, while the $\ddagger$ document-based methods.
    The best and second results in their own groups are marked in \textcolor{red}{\textbf{Red}} and \textcolor{blue}{\textbf{Blue}}, respectively.
}

 \footnotesize
	\begin{tabular*}{\textwidth}{@{\extracolsep\fill}c|c|c|ccc|ccccccccc}
 
	\hline
        \hline
	\multirow{3}{*}{Type} & \multirow{3}{*}{Method}  & \multirow{3}{*}{Venue} & \multicolumn{3}{c}{ZSL} & \multicolumn{9}{|c}{GZSL} \\
        \cline{4-15}
  
        & & & \multicolumn{1}{c}{AWA2} & \multicolumn{1}{c}{CUB} & \multicolumn{1}{c}{SUN} & \multicolumn{3}{|c}{AWA2} & \multicolumn{3}{|c}{CUB}  & \multicolumn{3}{|c}{SUN} \\
        \cline{4-15}
        
        & & & \multicolumn{1}{c}{T1} & \multicolumn{1}{c}{T1} & \multicolumn{1}{c}{T1} & \multicolumn{1}{|c}{U} & \multicolumn{1}{c}{S} & \multicolumn{1}{c}{H} & \multicolumn{1}{|c}{U} & \multicolumn{1}{c}{S} & \multicolumn{1}{c}{H} & \multicolumn{1}{|c}{U} & \multicolumn{1}{c}{S} & \multicolumn{1}{c}{H} \\
        
        \hline
        \multirow{7}{*}{E} & \multirow{1}{*}{APN${^\dagger}$} & NeurIPS20 & \multicolumn{1}{c}{68.4} & \multicolumn{1}{c}{72.0} & \multicolumn{1}{c}{61.6} & \multicolumn{1}{|c}{57.1} & \multicolumn{1}{c}{72.4} & \multicolumn{1}{c}{63.9} & \multicolumn{1}{|c}{65.3} & \multicolumn{1}{c}{69.3} & \multicolumn{1}{c}{67.2} & \multicolumn{1}{|c}{41.9} & \multicolumn{1}{c}{34.0} & \multicolumn{1}{c}{37.6} \\

        & \multirow{1}{*}{TransZero$^\dagger$} & AAAI22 & \multicolumn{1}{c}{70.1} & \multicolumn{1}{c}{\textcolor{blue}{\textbf{76.8}}} & \multicolumn{1}{c}{65.6} & \multicolumn{1}{|c}{61.3} & \multicolumn{1}{c}{82.3} & \multicolumn{1}{c}{70.2} & \multicolumn{1}{|c}{\textcolor{blue}{\textbf{69.3}}} & \multicolumn{1}{c}{68.3} & \multicolumn{1}{c}{68.8} & \multicolumn{1}{|c}{52.6} & \multicolumn{1}{c}{33.4} & \multicolumn{1}{c}{40.8} \\
        
        & \multirow{1}{*}{DUET$^\dagger$} & AAAI23  & \multicolumn{1}{c}{69.9} & \multicolumn{1}{c}{72.3} & \multicolumn{1}{c}{64.4} & \multicolumn{1}{|c}{63.7} & \multicolumn{1}{c}{\textcolor{red}{\textbf{84.7}}} & \multicolumn{1}{c}{72.7} & \multicolumn{1}{|c}{62.9} & \multicolumn{1}{c}{72.8} & \multicolumn{1}{c}{67.5} & \multicolumn{1}{|c}{45.7} & \multicolumn{1}{c}{45.8} & \multicolumn{1}{c}{45.8} \\

        % & \multirow{1}{*}{ReZSL$^\dagger$} & TIP23 & ViT & \multicolumn{1}{c}{69.3} & \multicolumn{1}{c}{77.6} & \multicolumn{1}{c}{69.8} & \multicolumn{1}{|c}{62.8} & \multicolumn{1}{c}{82.1} & \multicolumn{1}{c}{71.2} & \multicolumn{1}{|c}{70.5} & \multicolumn{1}{c}{77.7} & \multicolumn{1}{c}{73.9} & \multicolumn{1}{|c}{57.9} & \multicolumn{1}{c}{45.5} & \multicolumn{1}{c}{51.0} \\

        & \multirow{1}{*}{ZSLViT$^\dagger$} & CVPR24 & \multicolumn{1}{c}{70.7} & \multicolumn{1}{c}{\textcolor{red}{\textbf{78.9}}} & \multicolumn{1}{c}{\textcolor{blue}{\textbf{68.3}}} & \multicolumn{1}{|c}{66.1} & \multicolumn{1}{c}{\textcolor{blue}{\textbf{84.6}}} & \multicolumn{1}{c}{\textcolor{blue}{\textbf{74.2}}} & \multicolumn{1}{|c}{\textcolor{red}{\textbf{69.4}}} & \multicolumn{1}{c}{\textcolor{red}{\textbf{78.2}}} & \multicolumn{1}{c}{\textcolor{red}{\textbf{73.6}}} & \multicolumn{1}{|c}{45.9} & \multicolumn{1}{c}{\textcolor{red}{\textbf{48.4}}} & \multicolumn{1}{c}{47.3} \\
        
        % & \multirow{1}{*}{CLIP$^{*\ddagger}$} & ICML21 & \multicolumn{1}{c}{-} & \multicolumn{1}{c}{-} & \multicolumn{1}{c}{-} & \multicolumn{1}{|c}{-} & \multicolumn{1}{c}{-} & \multicolumn{1}{c}{-} & \multicolumn{1}{|c}{55.2} & \multicolumn{1}{c}{54.8} & \multicolumn{1}{c}{55.0} & \multicolumn{1}{|c}{-} & \multicolumn{1}{c}{-} & \multicolumn{1}{c}{-} \\

        % & \multirow{1}{*}{CoOp$^{*\ddagger}$} & IJCV22 & \multicolumn{1}{c}{-} & \multicolumn{1}{c}{-} & \multicolumn{1}{c}{-} & \multicolumn{1}{|c}{-} & \multicolumn{1}{c}{-} & \multicolumn{1}{c}{-} & \multicolumn{1}{|c}{49.2} & \multicolumn{1}{c}{63.8} & \multicolumn{1}{c}{55.6} & \multicolumn{1}{|c}{-} & \multicolumn{1}{c}{-} & \multicolumn{1}{c}{-} \\

        & \multirow{1}{*}{I2DFormer$^\ddagger$} & NeurIPS22 & \multicolumn{1}{c}{\textcolor{blue}{\textbf{76.4}}} & \multicolumn{1}{c}{45.4} & \multicolumn{1}{c}{-} & \multicolumn{1}{|c}{\textcolor{blue}{\textbf{66.8}}} & \multicolumn{1}{c}{76.8} & \multicolumn{1}{c}{71.5} & \multicolumn{1}{|c}{35.3} & \multicolumn{1}{c}{57.6} & \multicolumn{1}{c}{43.8} & \multicolumn{1}{|c}{-} & \multicolumn{1}{c}{-} & \multicolumn{1}{c}{-} \\

        & \multirow{1}{*}{I2MVFormer$^\ddagger$} & CVPR23 & \multicolumn{1}{c}{73.6} & \multicolumn{1}{c}{42.1} & \multicolumn{1}{c}{-} & \multicolumn{1}{|c}{66.6} & \multicolumn{1}{c}{82.9} & \multicolumn{1}{c}{73.8} & \multicolumn{1}{|c}{32.4} & \multicolumn{1}{c}{63.1} & \multicolumn{1}{c}{42.8} & \multicolumn{1}{|c}{-} & \multicolumn{1}{c}{-} & \multicolumn{1}{c}{-} \\

        & \multirow{1}{*}{\textbf{I2CFormer$^\dagger$} }  & \textbf{Ours} & \multicolumn{1}{c}{69.6} & \multicolumn{1}{c}{73.5} & \multicolumn{1}{c}{66.6} & \multicolumn{1}{|c}{61.4} & \multicolumn{1}{c}{83.9} & \multicolumn{1}{c}{70.9} & \multicolumn{1}{|c}{68.1} & \multicolumn{1}{c}{72.7} & \multicolumn{1}{c}{70.3} & \multicolumn{1}{|c}{\textcolor{blue}{\textbf{53.1}}} & \multicolumn{1}{c}{44.9} & \multicolumn{1}{c}{\textcolor{blue}{\textbf{48.7}}} \\
        
         & \multirow{1}{*}{\textbf{InfZSL+I2CFormer$^\dagger$} }  & \textbf{Ours} & \multicolumn{1}{c}{\textcolor{red}{\textbf{76.6}}} & \multicolumn{1}{c}{76.6} & \multicolumn{1}{c}{\textcolor{red}{\textbf{69.0}}} & \multicolumn{1}{|c}{\textcolor{red}{\textbf{69.3}}} & \multicolumn{1}{c}{83.6} & \multicolumn{1}{c}{\textcolor{red}{\textbf{75.8}}} & \multicolumn{1}{|c}{69.0} & \multicolumn{1}{c}{\textcolor{blue}{\textbf{74.5}}} & \multicolumn{1}{c}{\textcolor{blue}{\textbf{71.6}}} & \multicolumn{1}{|c}{\textcolor{red}{\textbf{54.7}}} & \multicolumn{1}{c}{\textcolor{blue}{\textbf{44.5}}} & \multicolumn{1}{c}{\textcolor{red}{\textbf{49.1}}} \\
         
         \hline
        
       \multirow{6}{*}{G}    & \multirow{1}{*}{HSVA$^{\dagger}$} & NeurIPS21 & \multicolumn{1}{c}{-} & \multicolumn{1}{c}{62.8} & \multicolumn{1}{c}{63.8} & \multicolumn{1}{|c}{59.3} & \multicolumn{1}{c}{76.6} & \multicolumn{1}{c}{66.8} & \multicolumn{1}{|c}{52.7} & \multicolumn{1}{c}{58.3} & \multicolumn{1}{c}{55.3} & \multicolumn{1}{|c}{48.6} & \multicolumn{1}{c}{39.0} & \multicolumn{1}{c}{43.3} \\

        & \multirow{1}{*}{DSP$^{\dagger}$} & ICML23 & \multicolumn{1}{c}{-} & \multicolumn{1}{c}{-} & \multicolumn{1}{c}{-} & \multicolumn{1}{|c}{60.0} & \multicolumn{1}{c}{86.0} & \multicolumn{1}{c}{70.7} & \multicolumn{1}{|c}{51.4} & \multicolumn{1}{c}{63.8} & \multicolumn{1}{c}{56.9} & \multicolumn{1}{|c}{48.3} & \multicolumn{1}{c}{43.0} & \multicolumn{1}{c}{45.5}  \\

        & \multirow{1}{*}{VADS$^{\dagger}$} & CVPR24 & \multicolumn{1}{c}{82.5} & \multicolumn{1}{c}{86.8} & \multicolumn{1}{c}{76.3} & \multicolumn{1}{|c}{\textcolor{red}{\textbf{75.4}}} & \multicolumn{1}{c}{83.6} & \multicolumn{1}{c}{79.3} & \multicolumn{1}{|c}{74.1} & \multicolumn{1}{c}{74.6} & \multicolumn{1}{c}{74.3} & \multicolumn{1}{|c}{\textcolor{red}{\textbf{64.6}}} & \multicolumn{1}{c}{49.0} & \multicolumn{1}{c}{55.7}  \\

        & \multirow{1}{*}{ZeroDiff$^{\dagger}$}  & ICLR25 &  \multicolumn{1}{c}{\textcolor{blue}{\textbf{87.3}}} &  \multicolumn{1}{c}{\textcolor{blue}{\textbf{87.5}}} & \multicolumn{1}{c}{\textcolor{blue}{\textbf{77.3}}} & \multicolumn{1}{|c}{74.7} &  \multicolumn{1}{c}{\textcolor{blue}{\textbf{89.3}}} & \multicolumn{1}{c}{\textcolor{blue}{\textbf{81.4}}} & \multicolumn{1}{|c}{\textcolor{blue}{\textbf{80.0}}}  & \multicolumn{1}{c}{\textcolor{blue}{\textbf{83.2}}} &  \multicolumn{1}{c}{\textcolor{blue}{\textbf{81.6}}} & \multicolumn{1}{|c}{63.0} &  \multicolumn{1}{c}{\textcolor{blue}{\textbf{56.9}}} & \multicolumn{1}{c}{\textcolor{blue}{\textbf{59.8}}}  \\

        & \multirow{1}{*}{\textbf{InfZSL+ZeroDiff$^{\dagger}$}} & \textbf{Ours} & \multicolumn{1}{c}{\textcolor{red}{\textbf{88.0}}} & \multicolumn{1}{c}{\textcolor{red}{\textbf{87.9}}} & \multicolumn{1}{c}{\textcolor{red}{\textbf{77.7}}} & \multicolumn{1}{|c}{\textcolor{blue}{\textbf{75.1}}} & \multicolumn{1}{c}{\textcolor{red}{\textbf{89.4}}} & \multicolumn{1}{c}{\textcolor{red}{\textbf{81.6}}} & \multicolumn{1}{|c}{\textcolor{red}{\textbf{81.2}}} & \multicolumn{1}{c}{\textcolor{red}{\textbf{83.4}}} & \multicolumn{1}{c}{\textcolor{red}{\textbf{82.3}}} & \multicolumn{1}{|c}{\textcolor{blue}{\textbf{63.3}}} & \multicolumn{1}{c}{\textcolor{red}{\textbf{58.2}}} & \multicolumn{1}{c}{\textcolor{red}{\textbf{60.7}}} \\

        \hline
        \hline
	
        \end{tabular*}
	\label{table:OverallComparsion}

\end{table*}

The results are reported in Table~\ref{table:OverallComparsion}.
We highlight two main observations:
\begin{enumerate}
    \item Generally, concept-based methods are better  than document-based methods in coarse-grained AWA2, but worse in fine-grained CUB.
    Our InfZSL fills the gap in AWA2 and keeps the performance in CUB.
    In AWA2, our InfZSL excels I2DFormer 3.3\% H and exceeds I2MVFormer 2.0 \% H for GZSL.
    In CUB, our InfZSL still has the second best results.
    \item Compared to all concept-based (embedding and generative) methods, InfZSL achieves the best performance AWA2 and SUN datasets
    For ZSL, our InfZSL has the best performance 76.5\%, and 69.0\% on AWA2 and SUN for ZSL, respectively.
    These are significant boosts compared to other concept-based embedding methods, i.e., by 5.8\% and 0.7\% on AWA2 and SUN, respectively.
    For GZSL, our InfZSL also has 1.6\% H and 1.8\% H improvements.
\end{enumerate}

\subsection{Ablation study}
\label{sec:ablation}

\subsubsection{Component Effectiveness}
\label{sec:Comp}
We examine the effect of the proposed components in terms of using the LLM-concept-based semantic embedding (denoted as `Inf.'), selecting concepts by transferability (denoted as `Tran.'), selecting concepts by discriminability (denoted as `Dis.') and the concept attention module  (denoted as `Att.') in our I2CFormer.
 Results are shown in Table~\ref{table:comp} in \textbf{Appendix \ref{app:addAbl}}.
We observe that `Inf.'+`Dis.' and `Inf.'+`Tran.' consistently both improve the performance across the three datasets, but `Inf.'+`Tran.' generally have large improvements.
It indicates  transferable concepts are more important than discriminative concepts in ZSL.
Combining `Dis.' and `Tran.', the performances increase further, reflecting the effectiveness of our harmonizing selection strategy.
Finally, the `Att.' module enhances our model by a large span, showing the importance of concept-specific attention.

\subsubsection{Hyper-parameters Analysis}
\label{sec:para}
\begin{figure}[htbp]
\centering
\begin{minipage}[t]{0.48\textwidth}
    \centering
    \includegraphics[width=\linewidth]{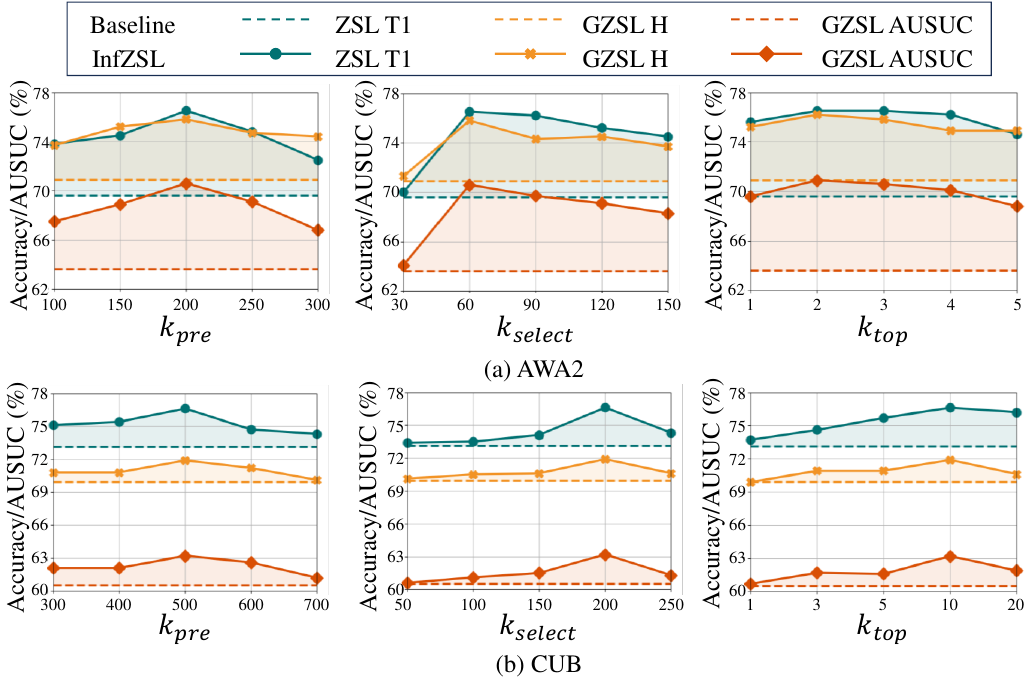}
    \caption{Hyper-parameters sensitivity analysis on (a) AWA2 and (b) CUB. The shaded area indicates the performance improvement compared to baseline.}
    \label{fig:parameters}
\end{minipage}
\hspace{.15in}
\begin{minipage}[t]{0.48\textwidth}
    \centering
    \includegraphics[width=\linewidth]{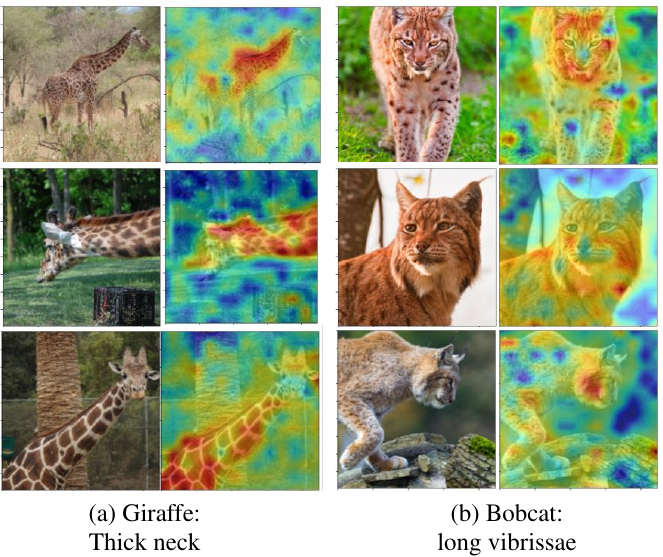}
    \caption{Attention visualizations of our LLM-generated concepts for two unseen classes.}
    \label{fig:attMap}
\end{minipage}
\end{figure}

We also perform hyper-parameter sensitivity analysis to investigate the effects of the cluster number $k_{pre}$, concept selection number $k_{select}$ and the parameter $k_{top}$ in Eq.~\ref{eq:dis} for measuring concept discriminability on Fig.~\ref{fig:parameters}.
We can find results increase gradually but decrease if the hyper-parameters are over-large.
It demonstrates the necessity of our hyper-parameter design.
And we provide our explain here.
Over-large $k_{pre}$ might over-separate similar concepts.
Over-large $k_{select}$ selects inappropriate concepts, e.g. good at only transferability or discriminability.
We also provide a class-concept heatmap visualization to illustrate $k_{select}$ is a key parameter.
Over-large $k_{top}$ leads to conditional probability $p(y|\mathbf{q})$ of concept discriminability (Eq.~\ref{eq:dis}) falling into unrelated classes.

\subsection{Qualitative Results}
\label{sec:vis}

\textbf{LLM-generated Concepts}
We provide the results about generated concepts and corresponding document-version in \textbf{Appendix \ref{app:addLLM}}.
The results exhibit that, even we explicitly prompt that we need visual semantics, LLMs still produce output containing non-visual semantics.

\textbf{Semantic Embedding Visualization}
To further verify our method can mine automatically trustworthy class concepts, we provide the heatmap visualization of our LLM-concept-based sematic embedding.
We sort concepts by their concept importance $r_{\mathbf{q}}$.
More left, higher transferability\& discriminability.
We can find our method digs out many potential key concepts with correct class co-occurrences.
For example, the most important concept `small dorsal fin' highlights with killer whale, blue whale, dolphin and so on.
The second important concept `long vibrissae' highlights with beaver, mole, leopard and so on.

\textbf{Concept Attention Visualization} 
We provide the attention visualization of our InfZSL to LLM-generated concepts in Fig.~\ref{fig:attMap}.
We show two significant concepts to the two unseen classes giraffe and bobcat.
Our InfZSL has a desirable visual-semantic alignment to different concepts.
More results for human-annotated concepts and analysis can be found in \textbf{Appendix \ref{app:addAtt}}.

\textbf{Visualization the effects of $k_{top}$} We visualize the heatmap regarding to different $k_{top}$ in Fig.~\ref{fig:heatmap_change}.
We find that when $k_{top}$ is zero, the concepts in the right end is not discriminative for few classes.
But when we increase $k_{top}$, the over-discriminative concepts are filled out.
It verified that our concept selection strategy can select those concepts having both high discriminative and transferability.

\section{Conclusion}
In this work, we devise a novel interpretable zero-shot learning framework InfZSL to leverage LLM-generated infinite class concepts.
To automatize the class concept generation, selection and scoring, we design a new prompt template to extract any number of class concepts from LLMs and conduct a Concept-Entropy based concept Selection and Scoring (CESS) strategy.
It does not only eliminates the hallucinated non-visual concepts, but also effectively discovers essential visual concepts that both have high transferability and discriminability.
We quantitatively and qualitatively demonstrate that InfZSL achieves consistent improvements over the current SOTAs on three ZSL benchmarks.

\bibliography{infzsl_arxiv}
\bibliographystyle{plain}

%%%%%%%%%%%%%%%%%%%%%%%%%%%%%%%%%%%%%%%%%%%%%%%%%%%%%%%%%%%%

\appendix

\section{Additional Experiments}
\label{app:addExp}

\subsection{Additional Ablation Study}
\label{app:addAbl}
\begin{table*}[htbp]
	\centering
	\caption{Ablation studies for different components of InfZSL. The symbols `Inf.', `Trans.', `Dis.' and `Att.' indicate using the semantic embedding based on LLM-generated concepts, selecting concepts by transferability and selecting concepts by discriminability, and our concept attention module in our I2CFormer, respectively.
}
 \tiny
	\begin{tabular*}{\textwidth}{@{\extracolsep\fill}c|c|c|c|ccc|ccccccccc}
 
	\hline
        \hline
	\multirow{3}{*}{Inf.} & \multirow{3}{*}{Dis.} & \multirow{3}{*}{Tran.} & \multirow{3}{*}{Att.} & \multicolumn{3}{c}{ZSL} & \multicolumn{9}{|c}{GZSL} \\
        \cline{5-16}
  
        & & & & \multicolumn{1}{c}{AWA2} & \multicolumn{1}{c}{CUB} & \multicolumn{1}{c}{SUN} & \multicolumn{3}{|c}{AWA2} & \multicolumn{3}{|c}{CUB}  & \multicolumn{3}{|c}{SUN} \\
        \cline{5-16}
        
        & &  &  & \multicolumn{1}{c}{T1} & \multicolumn{1}{c}{T1} & \multicolumn{1}{c}{T1} & \multicolumn{1}{|c}{U} & \multicolumn{1}{c}{S} & \multicolumn{1}{c}{H} & \multicolumn{1}{|c}{U} & \multicolumn{1}{c}{S} & \multicolumn{1}{c}{H} & \multicolumn{1}{|c}{U} & \multicolumn{1}{c}{S} & \multicolumn{1}{c}{H} \\
         \hline
         
        \multirow{1}{*}{$\times$} & \multirow{1}{*}{$\times$} & \multirow{1}{*}{$\times$} & $\times$ & \multicolumn{1}{c}{69.4} & \multicolumn{1}{c}{73.1} & \multicolumn{1}{c}{64.5} & \multicolumn{1}{|c}{61.4} & \multicolumn{1}{c}{83.9} & \multicolumn{1}{c}{70.9} & \multicolumn{1}{|c}{67.2} & \multicolumn{1}{c}{72.8} & \multicolumn{1}{c}{69.9} & \multicolumn{1}{|c}{53.8} & \multicolumn{1}{c}{43.9} & \multicolumn{1}{c}{48.4} \\
        
        \multirow{1}{*}{$\checkmark$} & \multirow{1}{*}{$\checkmark$} & \multirow{1}{*}{$\times$} & $\times$ & \multicolumn{1}{c}{70.1} & \multicolumn{1}{c}{74.1} & \multicolumn{1}{c}{67.2} & \multicolumn{1}{|c}{63.4} & \multicolumn{1}{c}{82.1} & \multicolumn{1}{c}{71.5} & \multicolumn{1}{|c}{65.8} & \multicolumn{1}{c}{75.9} & \multicolumn{1}{c}{70.5} & \multicolumn{1}{|c}{53.8} & \multicolumn{1}{c}{44.6} & \multicolumn{1}{c}{48.8} \\
        
        \multirow{1}{*}{$\checkmark$} & \multirow{1}{*}{$\times$} & \multirow{1}{*}{$\checkmark$} & $\times$ & \multicolumn{1}{c}{71.1} & \multicolumn{1}{c}{75.1} & \multicolumn{1}{c}{67.2} & \multicolumn{1}{|c}{65.3} & \multicolumn{1}{c}{83.7} & \multicolumn{1}{c}{73.3} & \multicolumn{1}{|c}{66.5} & \multicolumn{1}{c}{76.2} & \multicolumn{1}{c}{71.0} & \multicolumn{1}{|c}{53.8} & \multicolumn{1}{c}{45.1} & \multicolumn{1}{c}{49.0} \\

        \multirow{1}{*}{$\checkmark$} & \multirow{1}{*}{$\checkmark$} & \multirow{1}{*}{$\checkmark$} & $\times$ & \multicolumn{1}{c}{73.0} & \multicolumn{1}{c}{75.4} & \multicolumn{1}{c}{67.0} & \multicolumn{1}{|c}{63.6} & \multicolumn{1}{c}{87.0} & \multicolumn{1}{c}{73.5} & \multicolumn{1}{|c}{66.3} & \multicolumn{1}{c}{76.3} & \multicolumn{1}{c}{71.0} & \multicolumn{1}{|c}{53.1} & \multicolumn{1}{c}{44.9} & \multicolumn{1}{c}{48.7} \\

        \multirow{1}{*}{$\checkmark$}  & \multirow{1}{*}{$\checkmark$} & \multirow{1}{*}{$\checkmark$} & $\checkmark$ & \multicolumn{1}{c}{76.6} & \multicolumn{1}{c}{76.6} & \multicolumn{1}{c}{69.0} & \multicolumn{1}{|c}{69.3} & \multicolumn{1}{c}{83.6} & \multicolumn{1}{c}{75.8} & \multicolumn{1}{|c}{69.0} & \multicolumn{1}{c}{74.5} & \multicolumn{1}{c}{71.6} & \multicolumn{1}{|c}{54.7} & \multicolumn{1}{c}{44.5} & \multicolumn{1}{c}{49.1} \\
        \hline
        \hline
	
        \end{tabular*}
	\label{table:comp}
\end{table*}

\subsection{LLMs}
\label{app:addLLM}

\textbf{LLM-generated concepts}

\begin{figure*}[htbp]
    \centering
    \includegraphics[width=\linewidth]{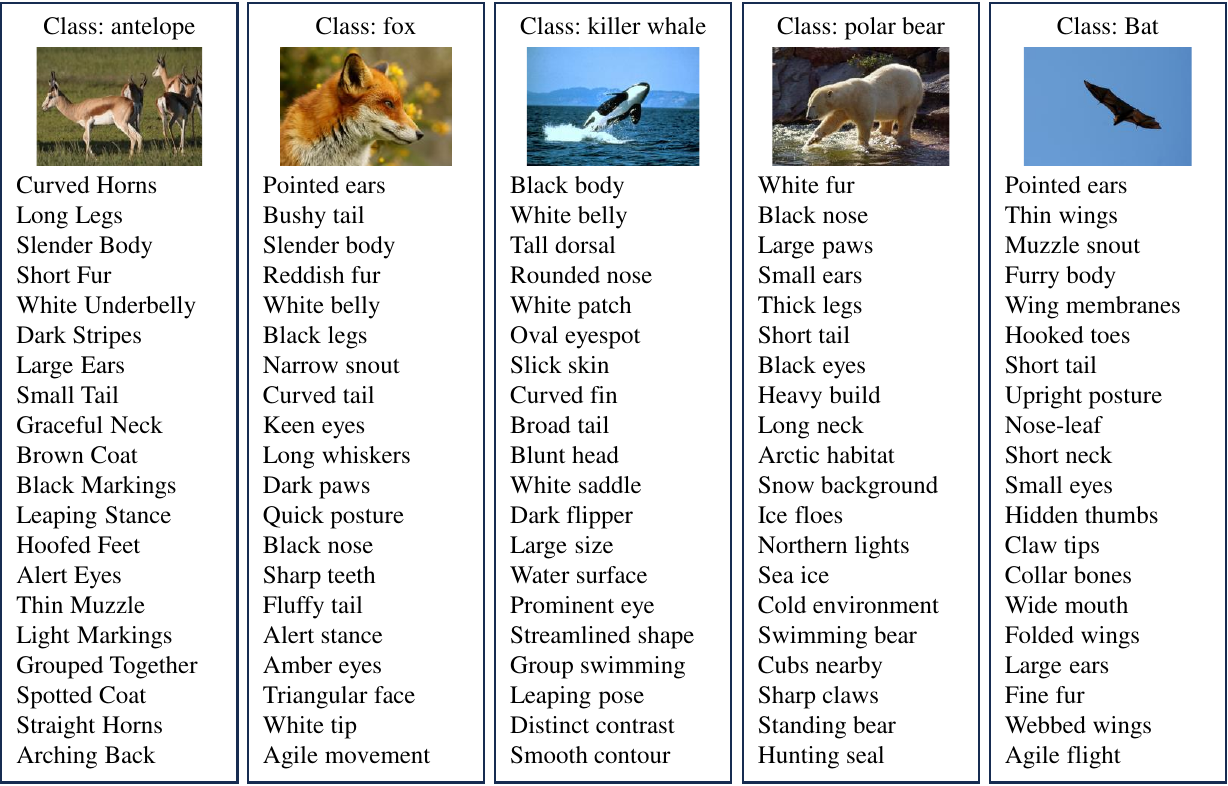}
    \caption{The examples of LLM-generated concepts on AWA2.}
    \label{fig:llm_con_awa2}
\end{figure*}

\begin{figure*}[htbp]
    \centering
    \includegraphics[width=\linewidth]{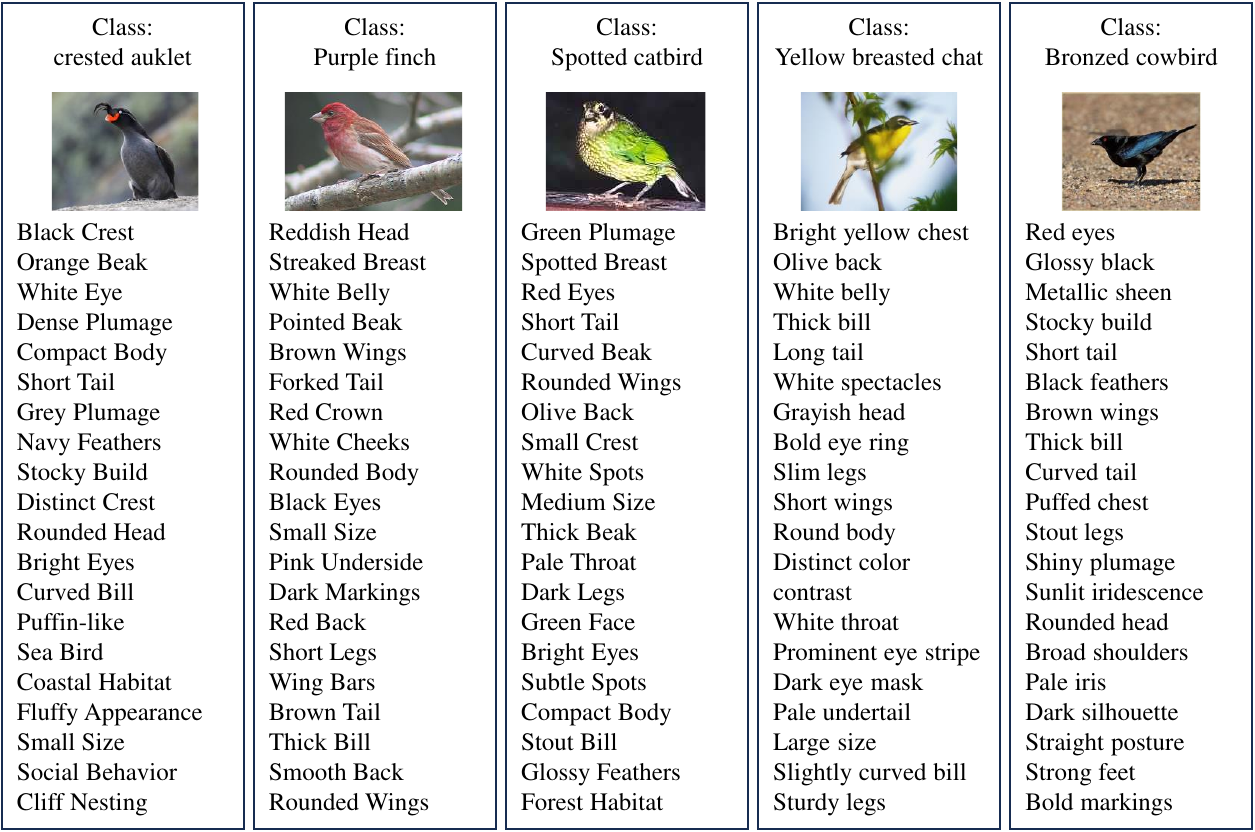}
    \caption{The examples of LLM-generated concepts on CUB.}
    \label{fig:llm_con_cub}
\end{figure*}

We provide the concept examples generated by LLMs in Fig.~\ref{fig:llm_con_awa2} for AWA2 and Fig.~\ref{fig:llm_con_cub} for CUB.

\textbf{LLM-generated documents}

\begin{figure*}[htbp]
    \centering
    \includegraphics[width=\linewidth]{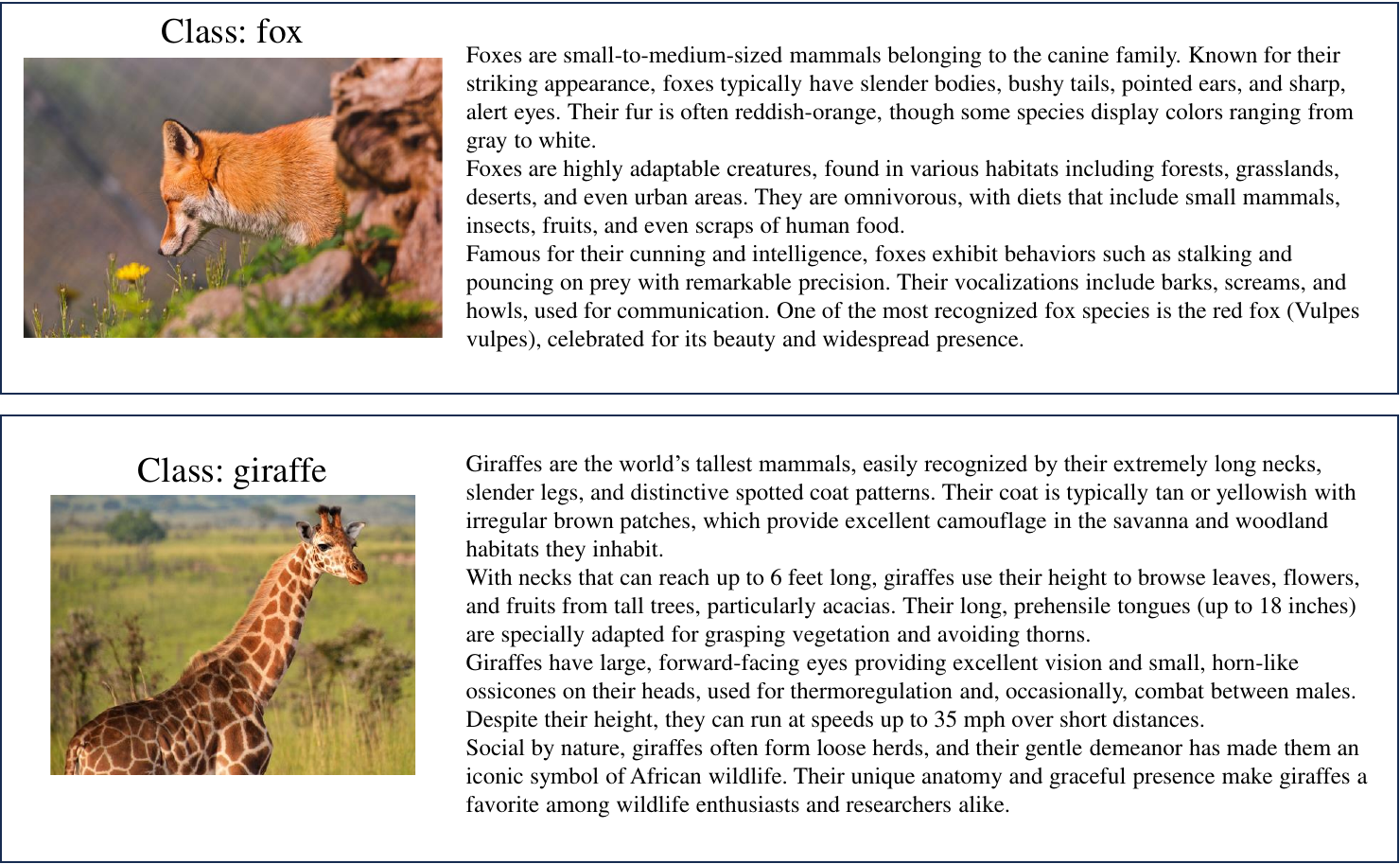}
    \caption{The examples of LLM-generated documents on AWA2.}
    \label{fig:llm_doc_awa2}
\end{figure*}

We use the prompt of I2MVformer~\cite{naeem2023i2mvformer} for prompting LLMs generate class documents.

\textit{
``A person wants to recognize $\{class\_type\}$ in images. They come across $\{class\_name\}$ and search online for facts about $\{class\_name\}$. They think the following description of $\{class\_name\}$ is a good description."
}

Two examples are exhibited in Fig.~\ref{fig:llm_doc_awa2}.
We can find even we have clearly we need facts about images. The generated documents still contain many non-visual descriptions.
It brings potential risk of document-based ZSL methods.

\subsection{Visualization the effects of selection parameter}
\label{app:addTopk}
\begin{figure*}[htbp]
    \centering
    \includegraphics[width=\linewidth]{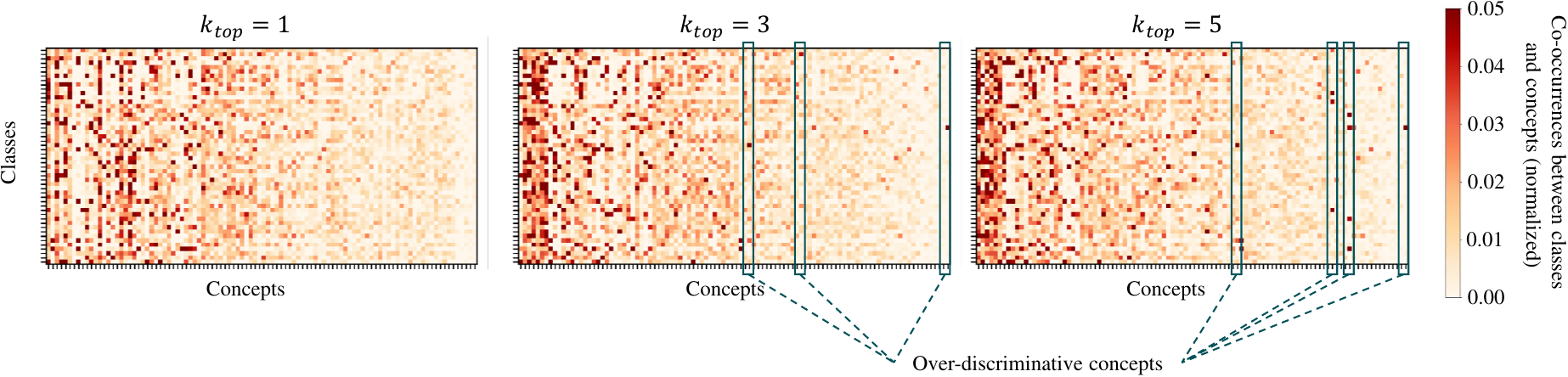}
    \caption{The effect of varying $k_{top}$.}
    \label{fig:heatmap_change}
\end{figure*}

We visualize the heatmap regarding to different $k_{top}$ in Fig.~\ref{fig:heatmap_change}.
We find that when $k_{top}$ is zero, the concepts in the right end is not discriminative for few classes.
But when we increase $k_{top}$, the over-discriminative concepts are filled out.
It verified that our concept selection strategy can select those concepts having both high discriminative and transferability.

\subsection{Attention of human-annotated concepts}
\label{app:addAtt}
\begin{figure*}[htbp]
    \centering
    \includegraphics[width=\linewidth]{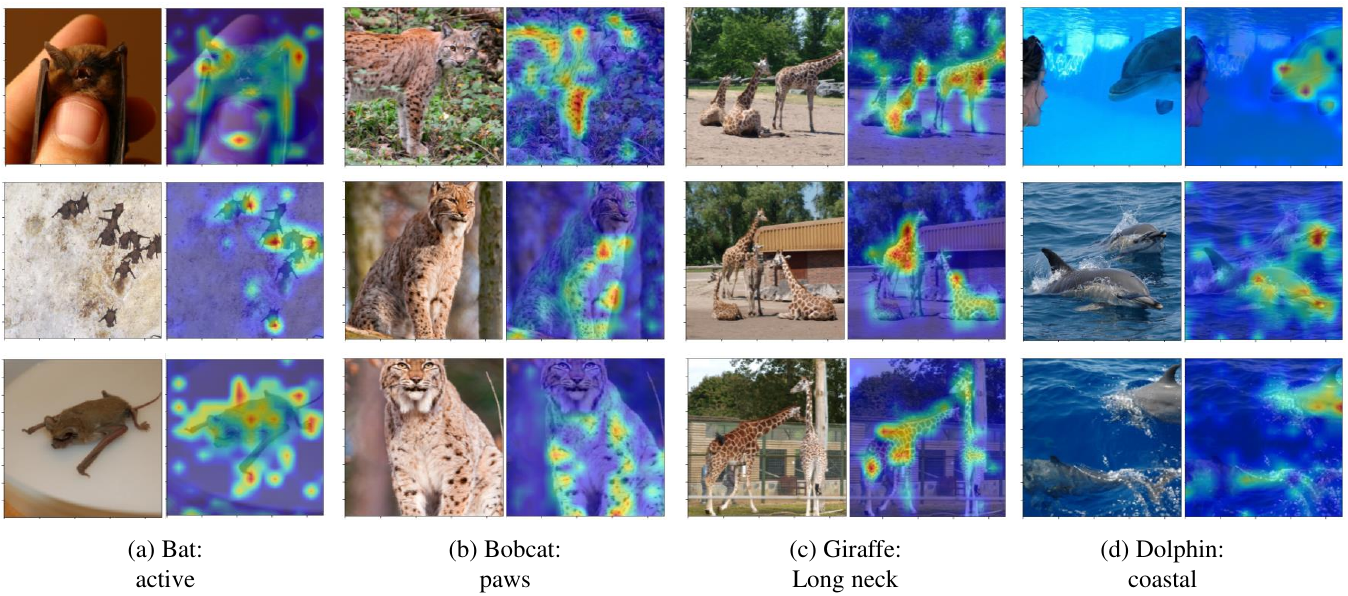}
    \caption{Attention visualizations of human-annotated concepts for four unseen classes.}
    \label{fig:attMap_h}
\end{figure*}

We visualize the attention map for human-annotated concepts in Fig.~\ref{fig:attMap_h}.
It shows our method also can correctly focus on related regions for human-annotated concepts.

\subsection{User Study}
\label{app:UserStudy}

\begin{figure*}[htbp]
    \centering
    \includegraphics[width=\linewidth]{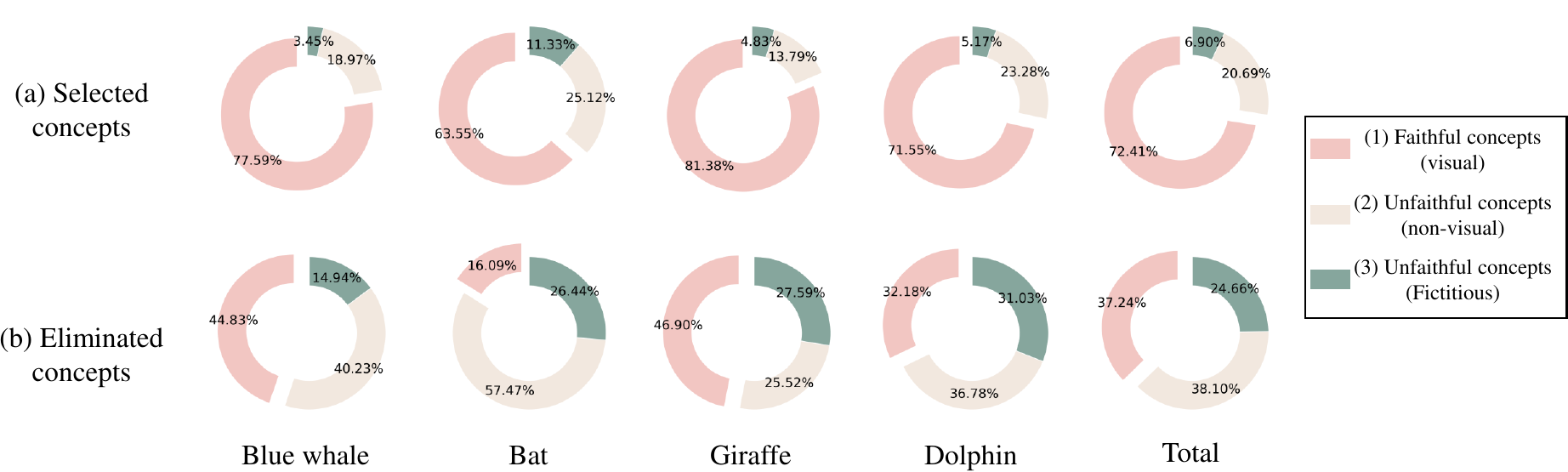}
    \caption{User study to our (a) selected and (b) eliminated concepts.}
    \label{fig:userStudy}
\end{figure*}

We select 4 unseen classes and 10 LLM-generated concepts randomly from AWA2.
Then we invited 29 volunteers to vote these concepts into three types: (1) Visual concepts (faithful); (2) Non-visual concepts
(Unfaithful) and (3) Fictitious concepts (Unfaithful). Finally, we count the ratios of our selected and eliminated concepts are voted into the
three types.
The result is shown in Fig.~\ref{fig:userStudy}.
It shows most of our selected concepts are visual concepts and eliminated concepts mainly are unfaithful. It further verifies the effectiveness of our method.

\section{I2CFormer}
\label{app:I2CFormer}

\begin{figure*}[htbp]
    \centering
    \includegraphics[width=\linewidth]{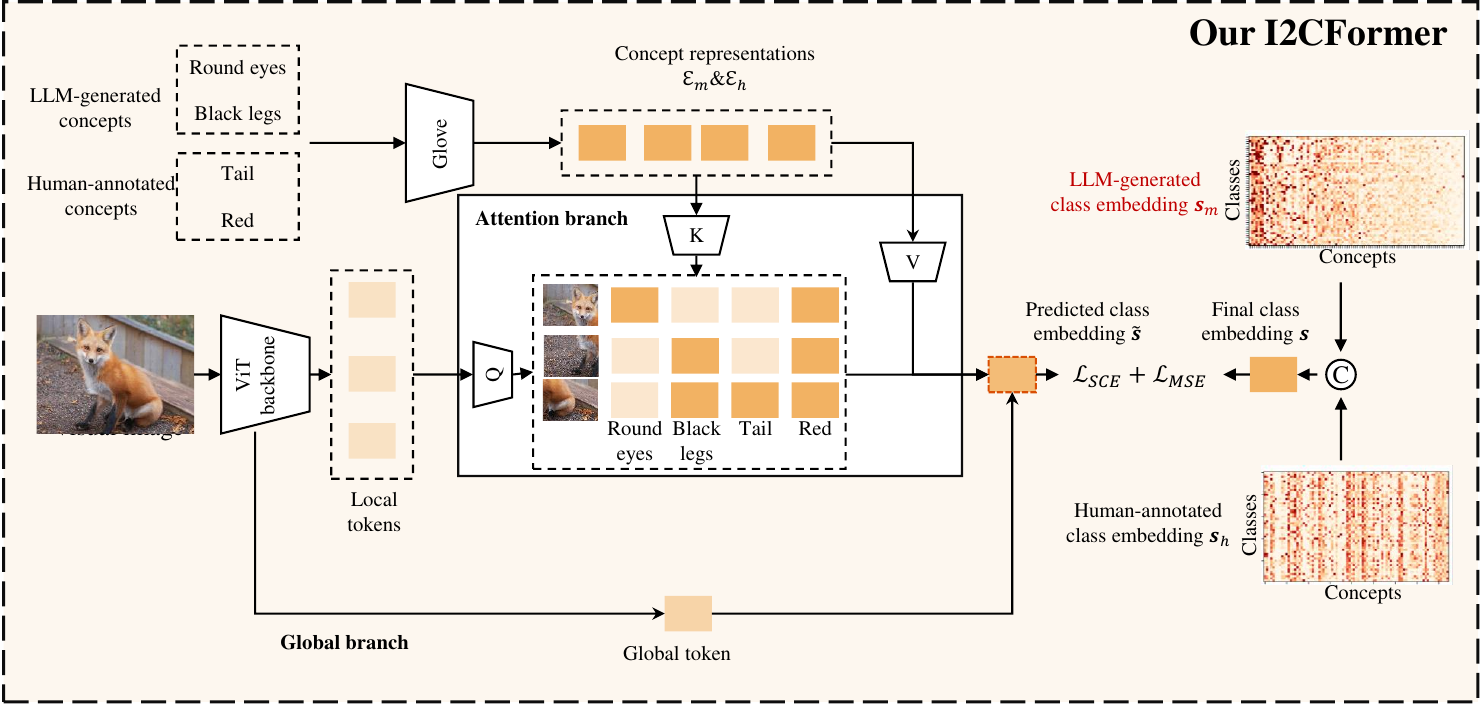}
    \caption{The detailed architecture of our I2DFormer.}
    \label{fig:i2cformer}
\end{figure*}

\textbf{Overall.} Following document-based ZSL work I2DFormer~\cite{naeem2022i2dformer}, we propose I2CFormer framework that consists of a global branch and an I2C attention module as shown in Fig.~\ref{fig:i2cformer}.
The global branch directly predict class embedding from the global feature of input image $\mathbf{x}$.
The I2C attention module also takes GloVe representations of human-annotated concepts $\mathcal{E}_{h}$ and selected LLM-generated ones $\mathcal{E}_{m}^{\prime}$ to produce concept-specific visual features.
In other word, the predicted class embedding is $\tilde{\mathbf{s}} = I2CFormer(\mathbf{x}, \mathcal{E}_{m}^{\prime}, \mathcal{E}_{h})$.

\textbf{Optimizing and Implementation.}
Our I2CFormer is trained by the widely-used Semantic Cross Entropy loss~\cite{xu2020attribute,chen2022transzero} ($\mathcal{L}_{SCE}$) and Mean-Squared Error loss~\cite{liu2021goal,ye2023rebalanced} $\mathcal{L}_{MSE}$.
The batch size is set to 32.
Our I2CFormer is optimized by Adam optimizer with a learning rate of 0.0005, momentum of 0.9, and weight decay of 0.0001.

\textbf{Inference.} During training, the model merely learns about the knowledge of seen categories, whereas both seen and unseen categories are available at inference time.
\begin{align}
    \tilde{y} = \arg \max_{y\in \mathcal{Y}^{u}} \cos(\mathbf{s}_{y},\tilde{\mathbf{s}}).
\end{align}
In the GZSL setting, seen class images also may be taken for testing, in which $\mathcal{X}^{u}$ and $\mathcal{Y}^{u}$ will be replaced by $\mathcal{X}^{s} \cup \mathcal{X}^{u}$ and $\mathcal{Y}^{s} \cup \mathcal{Y}^{u}$, respectively.

\subsection{Data-efficient ZSL}
\label{app:DEZSL}

\begin{table*}[htbp]
\setlength\tabcolsep{2pt}
	\centering
        \small
	\caption{Comparison on limited training data. We evaluate generative methods with 30\% and 10\% training samples. \( T1 \) represents the top-1 accuracy (\%) of unseen classes in ZSL. In GZSL, \( H \) represent the harmonic mean for top-1 accuracies on unseen classes and seen classes. The best and second-best results are marked in \textcolor{red}{\textbf{Red}} and \textcolor{blue}{\textbf{Blue}}, respectively.
}
	\begin{tabular*}{0.7\textwidth}{@{\extracolsep\fill}l |ccccc }

         \hline
         \hline
	\multirow{3}{*}{Method} & \multirow{3}{*}{Venue} & \multicolumn{4}{|c}{AWA2} \\
         \cline{3-6}

         & & \multicolumn{2}{|c}{$30\% \mathcal{D}^{tr}$} & \multicolumn{2}{|c}{$10\% \mathcal{D}^{tr}$} \\
         \cline{3-6}
         
         & & \multicolumn{1}{|c}{ZSL T1} & \multicolumn{1}{|c}{GZSL H} & \multicolumn{1}{|c}{ZSL T1} & \multicolumn{1}{|c}{GZSL H} \\
         \hline
         
         f-CLSWGAN & CVPR18 & \multicolumn{1}{|c}{68.9} & \multicolumn{1}{|c}{57.8} & \multicolumn{1}{|c}{54.0} & \multicolumn{1}{|c}{35.7} \\
         
        f-VAEGAN & CVPR19& \multicolumn{1}{|c}{81.2} & \multicolumn{1}{|c}{64.9} & \multicolumn{1}{|c}{73.1} & \multicolumn{1}{|c}{54.4}  \\
        
        \multirow{1}{*}{CEGAN} & CVPR21 & 
        \multicolumn{1}{|c}{72.2} & \multicolumn{1}{|c}{70.4} & \multicolumn{1}{|c}{69.0} & \multicolumn{1}{|c}{66.3} \\
        
         \multirow{1}{*}{DFCAFlow} & TCSVT23 & \multicolumn{1}{|c}{74.5} & \multicolumn{1}{|c}{72.6} & \multicolumn{1}{|c}{77.9} & \multicolumn{1}{|c}{70.7} \\
         
        \multirow{1}{*}{ZeroDiff } & ICLR25 & \multicolumn{1}{|c}{\textcolor{blue}{\textbf{84.9}}} & \multicolumn{1}{|c}{\textcolor{blue}{\textbf{80.2}}} & \multicolumn{1}{|c}{\textcolor{blue}{\textbf{83.3}}} & \multicolumn{1}{|c}{\textcolor{blue}{\textbf{77.0}}} \\
        
        \multirow{1}{*}{\textbf{ZeroDiff+InfZSL} } & \textbf{Ours} & \multicolumn{1}{|c}{\textcolor{red}{\textbf{85.5}}} & \multicolumn{1}{|c}{\textcolor{red}{\textbf{80.7}}} & \multicolumn{1}{|c}{\textcolor{red}{\textbf{85.2}}} & \multicolumn{1}{|c}{\textcolor{red}{\textbf{78.9}}} \\
        
        \hline
        \hline
        
	\end{tabular*}
	\label{table:DEZSL}
\end{table*}

%%%%%%%%%%%%%%%%%%%%%%%%%%%%%%%%%%%%%%%%%%%%%%%%%%%%%%%%%%%%

\end{document}